\documentclass[10pt,twocolumn,letterpaper]{article}

\usepackage{wacv}
\usepackage{times}
\usepackage{epsfig}
\usepackage{graphicx}
\usepackage{amsmath}
\usepackage{amssymb}
\usepackage{tabularx}
\usepackage{multirow}
\usepackage{array}
\usepackage{rotating}
\usepackage{lineno}

\def\wacvPaperID{1314} 

\wacvfinalcopy 

\ifwacvfinal
\def\assignedStartPage{1} 
\fi

\ifwacvfinal
\usepackage[breaklinks=true,bookmarks=false]{hyperref}
\else
\usepackage[pagebackref=true,breaklinks=true,colorlinks,bookmarks=false]{hyperref}
\fi

\ifwacvfinal
\else
\fi

\usepackage[font=small,labelfont=bf,tableposition=top]{caption}

\let\oldtwocolumn\twocolumn
\renewcommand\twocolumn[1][]{%
    \oldtwocolumn[{#1}{
    \begin{center}
           \includegraphics[width=\textwidth]{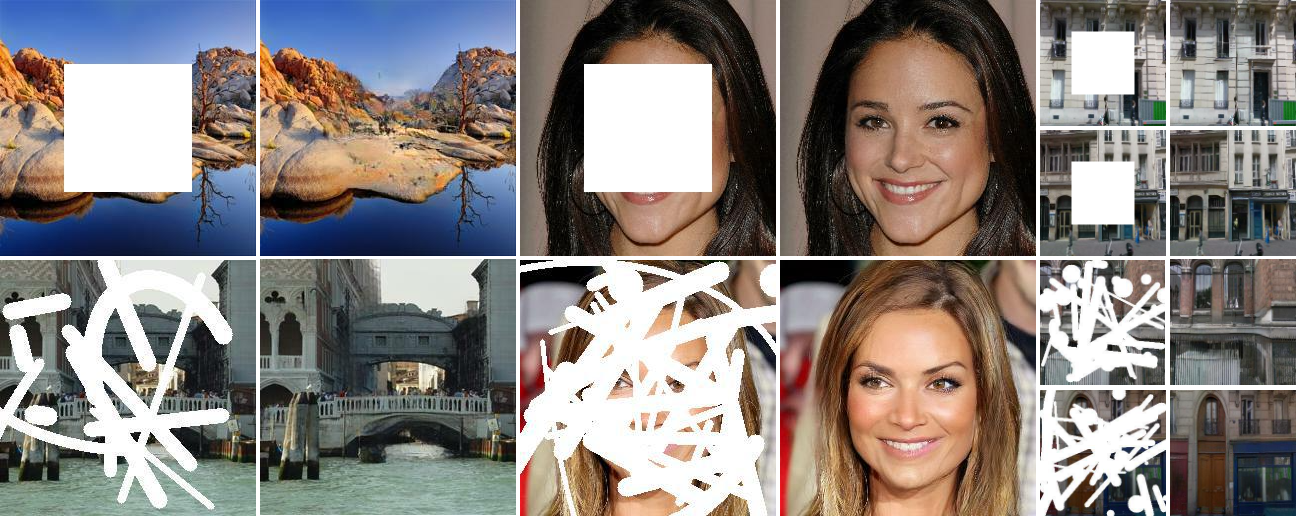} \vspace{-5mm}
           \captionof{figure}{Image Inpainting results by our method based on hypergraph convolution on spatial features. Each pair shows the input image and predicted image by our method. White pixels represent the missing data that needs to completed. [Best viewed in color]}
           \label{fig:fig1}
        \end{center}
    }]
}
\begin{document}

\title{Hyperrealistic Image Inpainting with Hypergraphs}

\author{Gourav Wadhwa$^{1}$  \hspace{3pt} Abhinav Dhall$^{1, 2}$ \hspace{3pt}  Subrahmanyam Murala$^{1}$ \hspace{3pt}  Usman Tariq$^{3}$\\
Indian Institute of Technology, Ropar$^{1}$ \hspace{3pt} Monash University$^{2}$ \hspace{3pt} American University of Sharjah$^{3}$\\
{\tt\small \{2017eeb1206, murala\}@iitrpr.ac.in \hspace{2pt} abhinav.dhall@monash.edu \hspace{2pt} utariq@aus.edu}}

\maketitle

\begin{abstract} {\vspace{-10mm}}
    Image inpainting is a non-trivial task in computer vision due to multiple possibilities for filling the missing data, which may be dependent on the global information of the image. Most of the existing approaches use the attention mechanism to learn the global context of the image. This attention mechanism produces semantically plausible but blurry results because of incapability to capture the global context. In this paper, we introduce hypergraph convolution on spatial features to learn the complex relationship among the data. We introduce a trainable mechanism to connect nodes using hyperedges for hypergraph convolution. To the best of our knowledge, hypergraph convolution have never been used on spatial features for any image-to-image tasks in computer vision.Further, we introduce gated convolution in the discriminator to enforce local consistency in the predicted image. The experiments on Places2, CelebA-HQ, Paris Street View, and Facades datasets, show that our approach achieves state-of-the-art results. 
\end{abstract}

\section{Introduction}
\setlength{\abovedisplayskip}{5pt}
\setlength{\belowdisplayskip}{5pt}
Image inpainting is the task of filling the missing regions such that modifications in the image are semantically plausible and can be further used in real-world applications such as restoring damaged or corrupted parts, removing distracting features from images, and completing occluded regions. There have been many learning and non-learning methods proposed in the past few decades. However, due to its inherent equivocalness and complexity in the natural images, image inpainting remains a challenging task.

To create a semantically plausible and realistic image, generally there are two requirements, $(a)$ global semantic structure, and $(b)$ fine detailed texture around the holes. Capturing of global semantic structure is non-trivial as a trained model can be easily biased towards producing blurred content.  Current image inpainting methods can be broadly divided into two categories: $1.$ content or texture copying approaches \cite{ref-102, ref-102, ref-104}, and $2.$ generative networks based approaches \cite{ref-13, ref-400, ref-403}.

The first method, content or texture copying, borrows the content or textures from the non-hole pixels to fill the missing regions. An example is total variations (TV) \cite{ref-100, ref-101} based approaches, which exploit the smoothness property in the image to fill in the missing regions. The patch matching approach borrows content from the surroundings to fill the missing regions. Patch Match algorithms \cite{ref-102, ref-103, ref-104, ref-105, ref-106} iteratively fill the missing pixels by searching the similar patches from the non-holes pixels in the image. These methods can effectively fill even the high-frequency missing content, however, are unable to identify the global semantic structure of the image producing improbable results.

Generative networks are being used in many computer vision tasks such as Image super-resolution \cite{ref-25, ref-26}, image de-blurring \cite{ref-27, ref-28}, image colorization \cite{ref-112, ref-113} etc. The generative network based approaches \cite{ref-114, ref-12, ref-116, ref-117, ref-118, ref-119, ref-13, ref-14} use these generative networks to predict the missing region in an image. These approaches learn to model distribution for the missing region conditioned on the image’s available surrounding regions. In \cite{ref-12}, the idea of using global and local discriminators to improve the local consistency of the completed image was proposed. These methods worked well when there are similar images in the training and test sets. However, these methods may not able to produce satisfying results for a totally different test image. Moreover, these methods produced artifacts for large irregular holes. In \cite{ref-116}, a patch swap mechanism between the Image2Feature network and Feature2Image network was used. This helped in combining the copying and deep learning approaches to map the uncompleted image with the completed images. \cite{ref-117, ref-118, ref-119} used novel contextual attention mechanisms to borrow the patches from a distant location.

Inspired by the \emph{hyperrealism art} genre of painting, which resembles the high-resolution images, we propose a novel image inpainting method using the hypergraphs structure. The proposed hypergraph structure enables the network to find matching features from the background to fill in the missing regions. We use a two-stage network (coarse and refine network) for image inpainting. Firstly, the coarse network roughly fills the missing region and then the refine network uses this coarse output to produce finer results. We introduce a novel data-dependent method for developing incidence matrix for hypergraph convolution. To the best of our knowledge, this is one of the first works to propose use of hypergraph convolution networks on spatial features for any image to image tasks in computer vision. We also show that our proposed method obtains substantially better results for both center and irregular mask for image inpainting. The proposed hypergraphs convolutional layer can easily be used for the other computer vision tasks such as image super-resolution, image de-blurring, to get a global context in the image. Further we introduce gated convolution in discriminator to enforce the local consistency in the predicted image. Our major contributions can be summarized as,

\begin{itemize}
    \item We propose a novel Image inpainting network using hypergraphs to produce globally semantic completed images. 
    \item We propose a trainable method to compute data-dependent incidence matrix for hypergraph convolutions.
    \item  We introduce gated convolution instead of regular convolutions in the discriminator, enabling it to enforce local consistency in the completed image.
\end{itemize}
Further, we train our network using a simple yet effective incremental strategy, which enables completion of the irregular holes. We also test our network on four publicly available datasets and show that our method performs significantly better than the previous state-of-the-art-methods.

\section{Related Work}
\textbf{Free Form Image Inpainting:} One of the major problem with CNNs for image inpainting is that they provide equal weight to each spatial pixel in the image and hence are unable to discriminate between the hole pixels and non-hole pixels. To go around this problem, \cite{ref-13} introduced a partial convolution that would allow different weightage for the hole and non-hole pixels. They applied a convolutional operation only on the hole pixels and then followed it by a rule-based update of the mask for the following layers. In \cite{ref-14}, the authors improved upon the idea of masked convolutions by introducing gated convolutions. Instead of the rule-based update of the mask, they introduced a trainable approach to find the mask values, where the masks are calculated using convolution operation and then multiplied by the spatial features to assign different weights for hole and non-hole pixels. In this work, we build upon this gated convolution framework to propose our method.

\begin{figure}[t] {\vspace{-4mm}}
\begin{center}
  \includegraphics[width=1.0\linewidth]{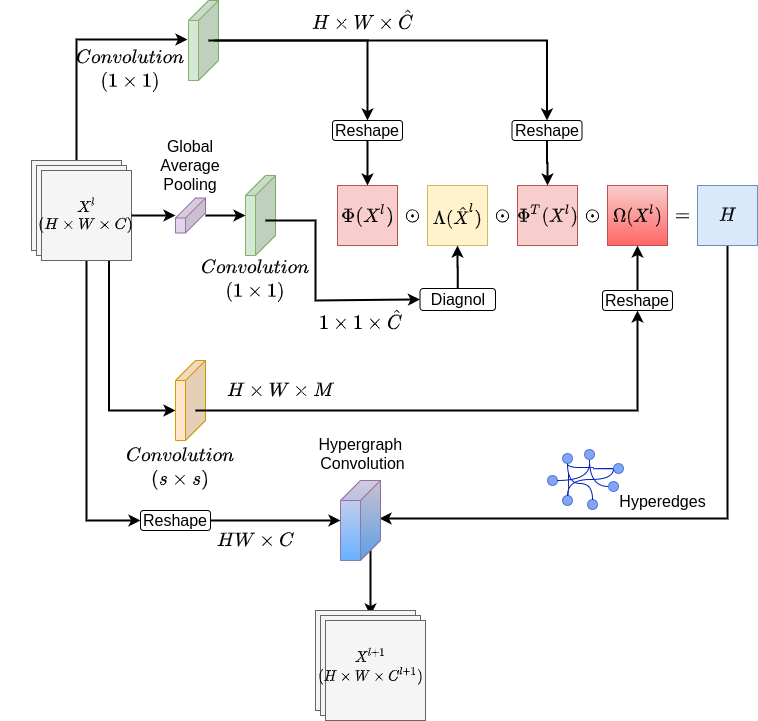}
\end{center}  \vspace{-5mm}
  \caption{Overview of our proposed hypergraph convolution on the spatial features. First, we compute the incidence matrix $H$ using the information gathered from the input features and then we compute the hypergraph convolution using the calculated incidence matrix as given by Eq - \eqref{eq-A7}.} \vspace{-2mm}
  \label{fig:fig2}
\end{figure}

\textbf{Graph Neural Networks (GNN):} Recently there has been a growing interest \cite{ref-201, ref-202, ref-203} to extend the deep learning approaches for the graph-related data. The conventional CNNs can be seen as a special case of graph data in which each spatial pixel is connected by its surrounding pixels. Graph neural networks can increase the network’s overall receptive field and hence enforce global consistency in the predictions \cite{ref-204}.  Despite the significant improvement in these methods, there has been a limited use of GNNs in image inpainting. GNNs have been used in some of the related fields such as image super-resolution \cite{ref-205}, semantic segmentation \cite{ref-206, ref-207}, image de-noising \cite{ref-208, ref-209} etc. \cite{ref-205} models the correlation between the cross-scale similar patches as a graph and introduce and patch aggregation module to build the high-resolution image from the low-resolution counterpart. In \cite{ref-208, ref-209}, the authors present a non-local aggregation block that uses a graph neural network for aggregating the features from far away pixels. To build the graph, they find the distance between the spatial features and chooses k-nearest neighbors. In \cite{ref-206}, the authors introduce a method for finding the similarity matrix for the graph, which is formed using each spatial pixel as vertex, using a data-dependent technique. They further use it in a pyramid based structure for the task of semantic segmentation. We extend this technique for the hypergraph neural networks to model a much more complex relation between the pixels using hyperedges, connecting more than two nodes using a single edge.

GNNs are efficient and can easily handle the long-range contextual information in the image, but they cannot accurately represent the non-pair relations among the data. Hypergraphs are a more generalized version of the graph in which a hyperedge can connect any number of vertices. Recently many researchers are using hypergraphs to represent their data in the deep learning approaches \cite{ref-210, ref-211}. In \cite{ref-2}, the authors proposed hypergraph neural network (HGNN) which introduces spectral convolution on hypergraphs, using the regularization framework introduced in \cite{ref-1}. In \cite{ref-6}, the authors introduce a hypergraph attention module. The hypergraph attention module further exerts an attention mechanism to learn the dynamic connections of hyperedges. Both of the previous methods cannot handle the dynamic structure of the hypergraphs. \cite{ref-215} introduces an idea of dynamic hypergraph construction, using the k-NN clustering method. This method is able to manage the dynamic nature of the input data, but it limits the number of nodes that can be connected. We propose a hypergraph inspired image inpainting method which can learn the hypergraph structure from the input data. 

\section{Methodology}
We start the method discussion with brief introduction to spectral convolution on hypergraphs and then present the details oftrainable hypergraph module (Figure \ref{fig:fig2}). Later, we discuss details of the inpainting network. 
\subsection{Hypergraph Convolution}
Hypergraphs structure is used in many computer vision tasks to model the high-order constraints on the data which cannot be accommodated in the conventional graph structure. Unlike the  pairwise connections in graphs, hypergraph contains hyperedge which can connect two or more vertices. A hypergraph is defined as $G = (V, E, \mathbf{W})$, where $V = \{v_{1}, \dots, v_{N}\}$ is the set of vertices, $E = \{e_{1}, \dots, e_{M}\}$ represents the set of hyperedges, and $\mathbf{W} \in \mathbb{R}^{M \times M}$ is a diagonal matrix containing the weight of each edge. The hypergraph $G$ can be represented by the incidence matrix $\mathbf{H} \in \mathbb{R}^{N \times M}$. For a vertex $v \in V$, and an edge $e \in E$ the incidence matrix is defined as,
\begin{equation}
    h (v, e) = 
    \left \{
    \begin{array}{ll}
          1  & \text{if }  v \in e  \\
          0 & \text {if }  v \not\in e \\
    \end{array} 
    \right. \label{eq-A1}
\end{equation}
For a given hypergraph $G$, the vertex degree, $\mathbf {D} \in \mathbb{R}^{N \times N}$, and hyperedge degree $\mathbf {B} \in \mathbb{R}^{M \times M}$ are defined as $D_{ii} = \sum_{e=1}^{M} W_{ee} H_{ie}$, and $B_{ee} = \sum_{i=1}^{N} H_{ie}$ respectively,
\\
\begin{figure*}
\begin{center}
\includegraphics[width=\linewidth]{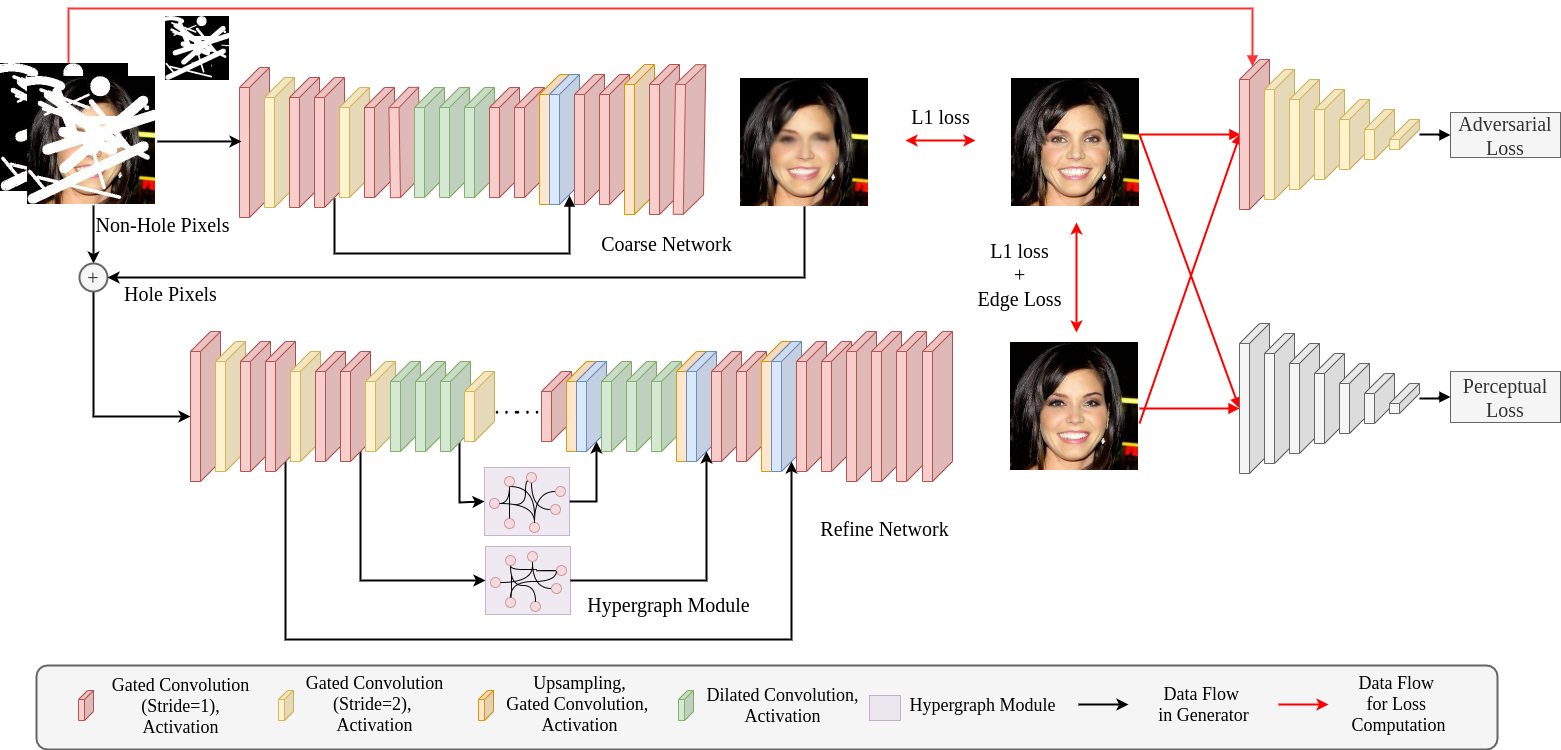}
\end{center}
 \vspace{-5mm}
  \caption{Overview of our proposed network for Image inpainting. The Coarse network roughly completes the missing holes. Later, the hypergraph convolution based Refine network generates the final high quality completed image.} \vspace{-2mm}
  \label{fig:fig3}
\end{figure*}

Next, the incidence matrix $H$, vertex degree $D$, and hyperedge degree $B$, are used to compute the normalized hypergraph Laplacian matrix $\mathbf{\Delta} \in \mathbb{R}^{N \times N}$ as, $\mathbf{\Delta} = \mathbf{I - D^{-1/2} H B^{-1} H^{T} D^{-1/2}}$. It is a symmetric positive semi-definite matrix \cite{ref-1} and the eigen decomposition $\mathbf{\Delta = \Phi \Lambda \Phi^{T}}$ can be used to get the complete set of the orthonormal eigenvectors $\mathbf{\Phi} = \{\phi_{1}, \dots, \phi_{N}\}$ and a diagonal matrix $\mathbf{\Lambda} = diag (\lambda_{1}, \dots, \lambda_{N})$ containing the corresponding non-negative eigenvalues. We can define the hypergraph Fourier Transform, $\mathbf{\hat{x} = \Phi^T x}$, which transforms a signal $\mathbf{x} = (x_{1}, \dots, x_{N})$ into the spectral domain spanned by the  basis of $\Phi$, also known as Fourier basis. Generalizing the convolutional theorem into structured space of hypergraphs, the convolution on the signal $x \in \mathbb{R}^{N}$ can be defined as:
\begin{equation}
    g \circledast x = \mathbf{\Phi} g(\mathbf{\Lambda}) \mathbf{\Phi^T} x \label{eq-A4}
\end{equation}
where $g(\mathbf{\Lambda}) = diag (g(\lambda_1), \dots, \lambda_{N})$ is a function of the Fourier coefficients \cite{ref-2}. However, to compute the convolution on the signal $x$ it would be required to compute the eigenvectors of the Laplacian matrix. So, Defferrand et al. \cite{ref-3} parameterized $g(\mathbf{\Lambda})$ with truncated chebyshev polynomials up to $K^{th}$ order, hence defining the convolutional operation on the hypergraph signal as,  
\begin{equation}
    g \circledast x = \sum_{k=0}^{K} \theta_{k} T_{k} (\mathbf {\Delta}) x \label{eq-A5}
\end{equation}
where $\theta_{k}$ is a vector of chebyshev polynomial coefficients, and $T_{k}$ is the chebyshev polynomial. In Eq - \eqref{eq-A5}, we excluded the calculation of eigenvectors of the Laplacian matrix. We can further simplify the formulation by limiting $K=1$. In \cite{ref-4}, it is approximated $\lambda_{max} \approx 2$ because of the scale adaptability of neural networks. Therefore, our convolutional operation on the hypergraph signal becomes,
\begin{equation}
    \begin{aligned}
        g \circledast x &\approx \theta \mathbf {D}^{-1/2} \mathbf{H} \mathbf{W} \mathbf{B}^{-1} \mathbf{H}^{T} \mathbf{D}^{-1/2} x \label{eq-A6}
    \end{aligned}
\end{equation}
where $\theta$ is the only chebyshev coefficient left after taking $K=1$ chebyshev polynomials. 

For a given hypergraph signal $X^{l} \in \mathbb{R}^{N \times C_{l}}$, where $C_{l}$ is the dimension of the feature vector of input at layer $l$, we can generalize the convolution operation in multi-layer hypergraph convolutional network as, 
\begin{equation}
    \mathbf{X^{l+1}} = \sigma (\mathbf {D}^{-1/2} \mathbf{H} \mathbf{W} \mathbf{B}^{-1} \mathbf{H}^{T} \mathbf{D}^{-1/2} \mathbf{X^{l}} \mathbf{\Theta}) \label{eq-A7}
\end{equation}
where $\Theta \in \mathbb{R}^{C_{l} \times C_{l+1}}$ is the learnable parameter, and $\sigma$ is the non-linear activation function.


In Eq  - \eqref{eq-A7}, the incidence matrix $H$ encodes the hypergraph structure, which is further used to propagate information among the hypergraph nodes. Hence, it can be easily seen that better hyperedges’ connections would lead to better information sharing among the nodes, further improving the completed image. Currently, the formation of these incidence matrices is limited to non-trainable methods.  

\subsection{Hypergraphs Convolution on spatial features}
To overcome the limited receptive field of CNN architectures, recent studies transform the spatial feature maps into the graph-based structure and perform graph convolution to capture the global relationship between the data \cite{ref-8, ref-9, ref-10}. It can be easily observed that simple graphs are a special case of hypergraphs where each hyperedge connects only two node. These simple graph can easily represent the pair-wise relationship among data but it is difficult to represent the complex relationship among the spatial features of the image because of which we use hypergraphs instead of graphs. To transform the spatial features $F^{l} \in \mathbb{R}^{h \times \times w \times c}$ into the graph-like structure, we consider each spatial feature as a node having a feature vector of dimension $c$, $X^{l} \in \mathbb{R}^{hw \times c}$. 
\begin{table*}[htbp]
    \centering
    \small
    \begin{tabularx}{\textwidth } {c | c || c | c | c | c | c || c | c | c | c}
          &  & \multicolumn{5}{c||}{CelebA-HQ} & \multicolumn{4}{c}{Places2} \\
          \hline
         \footnotesize{\%} & \footnotesize{Metrics} & \footnotesize{PICNet\cite{ref-403}} & \footnotesize{GMCNN\cite{ref-400}} & \footnotesize{DeepFill\cite{ref-14}} & \footnotesize{SN\cite{ref-117}} & \footnotesize{Ours} & \footnotesize{PICNet\cite{ref-403}} & \footnotesize{GMCNN\cite{ref-400}} & \footnotesize{DeepFill\cite{ref-14}} & \footnotesize{Ours} \\
         \hline
         \multirow{5}{*}{\begin{sideways}0.1-0.2 \end{sideways}}  & PSNR $\uparrow$ &  30.29 & 30.98 & 31.21 & 30.16 & \textbf{33.34} & 29.6 & 30.35 & 29.87 & \textbf{32.21} \\
         \cline {2-11}
         & SSIM $\uparrow$ & 0.971 & 0.977 & 0.9744 & 0.969 & \textbf{0.985} & 0.953 & 0.964 & 0.960 & \textbf{0.974} \\
         \cline {2-11}
         & FID $\downarrow$& 6.223 & 6.487 & 3.786 & 7.143 & \textbf{2.177} & 13.269 & 8.687 & 9.567 & \textbf{6.465} \\
         \cline{2-11}
         & L1 $\downarrow$& 2.004 & 1.9193 & 1.622 & 2.203 & \textbf{0.683} & 1.340 & 1.192 & 1.240 & \textbf{0.745} \\
         \cline{2-11}
         & L2 $\downarrow$& 0.234 & 0.203 & 0.187 & 0.235 & \textbf{0.124} & 0.309 & 0.273 & 0.306 & \textbf{0.184} \\
         \hline \hline
         \multirow{5}{*}{\begin{sideways}0.2-0.3 \end{sideways}} & PSNR $\uparrow$ & 28.10 & 28.84 & 28.52 & 28.55 & \textbf{30.23} & 26.54 & 27.35 & 26.89 & \textbf{29.13} \\
         \cline {2-11}
         & SSIM $\uparrow$& 0.951 & 0.961 & 0.955 & 0.954 & \textbf{0.970} & 0.911 & 0.932 & 0.924 & \textbf{0.950} \\
         \cline {2-11}
         & FID $\downarrow$& 8.343 & 8.931 & 6.013 & 9.342 & \textbf{4.026} & 21.496 & 14.250 & 15.007 & \textbf{11.175} \\
         \cline{2-11}
         & L1 $\downarrow$& 2.508 & 2.303 & 2.179 & 2.560 & \textbf{1.250} & 2.230 & 1.938 & 2.030 & \textbf{1.390} \\
         \cline{2-11}
         & L2 $\downarrow$& 0.375 & 0.329 & 0.339 & 0.339 & \textbf{0.240} & 0.603 &0.523& 0.588 & \textbf{0.357} \\
         \hline \hline
         \multirow{5}{*}{\begin{sideways}0.3-0.4 \end{sideways}} & PSNR $\uparrow$& 26.38 & 26.80 & 26.62 & 27.00 & \textbf{28.22} & 24.50 & 25.37 & 24.93 & \textbf{27.17} \\
         \cline {2-11}
         & SSIM $\uparrow$& 0.927 & 0.941 & 0.933 & 0.934 & \textbf{0.954} & 0.862 & 0.897 & 0.885 & \textbf{0.923} \\
         \cline {2-11}
         & FID $\downarrow$& 10.334 & 15.840 & 8.650 & 11.930 & \textbf{5.991} & 29.340 & 19.900 & 21.566 & \textbf{16.116} \\
         \cline{2-11}
         & L1 $\downarrow$& 3.070 & 2.819 & 2.784 & 3.047 & \textbf{1.834} & 3.190 & 2.710 & 2.850 & \textbf{2.048} \\
         \cline{2-11}
         & L2 $\downarrow$& 0.551 & 0.524 & 0.516 & 0.476 & \textbf{0.374} & 0.963 & 0.804 & 0.902 & \textbf{0.545} \\
         \hline \hline
         \multirow{5}{*}{\begin{sideways}0.4-0.5 \end{sideways}} & PSNR $\uparrow$& 24.92 & 24.49 & 25.08 & 25.39 & \textbf{26.69} & 22.95 & 23.79 & 23.40 &   \textbf{25.68} \\
         \cline {2-11}
         & SSIM $\uparrow$& 0.898 & 0.906 & 0.906 & 0.903 & \textbf{0.935} & 0.806 & 0.853 & 0.838 & \textbf{0.890} \\
         \cline {2-11}
         & FID $\downarrow$& 13.015 & 33.358 & 11.410 & 15.960 & \textbf{7.942} & 37.399 & 25.589 & 27.624 & \textbf{21.211} \\
         \cline{2-11}
         & L1 $\downarrow$& 3.730 & 3.588 & 3.469 & 3.827 & \textbf{2.454} & 4.218 & 3.574 & 3.750 & \textbf{2.750} \\
         \cline{2-11}
         & L2 $\downarrow$& 0.773 & 0.914 & 0.738 & 0.690 & \textbf{0.530} & 1.360 & 1.148 & 1.260 & \textbf{0.760} \\
         \hline \hline
         \multirow{5}{*}{\begin{sideways}0.5-0.6 \end{sideways}} & PSNR $\uparrow$& 23.47 & 21.33 & 23.67 & 22.90 & \textbf{25.27} & 21.51 & 22.37 & 22.08 & \textbf{24.35} \\
         \cline {2-11}
         & SSIM $\uparrow$& 0.86 & 0.842 & 0.873 & 0.838 & \textbf{0.911} & 0.742 & 0.802 & 0.785 & \textbf{0.851} \\
         \cline {2-11}
         & FID $\downarrow$ & 16.13 & 64.449 & 14.852 & 25.440 & \textbf{10.087} & 45.630 & 33.239 & 34.387 & \textbf{28.237} \\
         \cline{2-11}
         & L1 $\downarrow$ & 4.529 & 5.120 & 4.270 & 5.530 & \textbf{3.146} & 5.380 & 4.559 & 4.746 & \textbf{3.530} \\
         \cline{2-11}
         & L2 $\downarrow$ & 1.074 & 2.008 & 1.010 & 1.245 & \textbf{0.730} & 1.900 & 1.580 & 1.710 & \textbf{1.030} \\
         \hline \hline
    \end{tabularx} 
    \caption{Quantitative comparison of our method with state-of-the-art methods on CelebA-HQ and Places2 datasets wrt different hole percentages. ($\uparrow$ Higher is better, $\downarrow$ Lower is better)} \vspace{-8mm}
    \label{tab:my_label}
\end{table*}

In the recent studies \cite{ref-2, ref-6, ref-7}, for the visual classification problem, the incidence matrix $H$ is formed using the euclidean distance between features of the images \cite{ref-2, ref-7}. To better capture the image’s intra-spatial structure, we propose an improved incidence matrix that can learn to capture long-term intra-spatial dependencies. Instead of the euclidean distance between the spatial features, we use the cross correlation of the spatial features to calculate each node’s contribution in each hyperedge. The incidence matrix $H$ contains the information regarding each node’s contribution in each hyperedge and it is expressed as,
\begin{equation}
    H = \Psi(X)  \Lambda(X) \Psi(X)^{T} \Omega (X)
\end{equation}
where $\Psi(X) \in \mathbb{R}^{N \times \hat{C}}$, is the linear embedding of the input features followed by a non-linear activation function (in our case ReLU function), and $\hat{C}$ is the dimension of the feature vector after the linear embedding, $\Lambda(X) \in \mathbb{R}^{\hat{C} \times \hat{C}}$ is a diagonal matrix, which helps in learning a better distance metric among the nodes for the incidence matrix $H$, and $\Omega (X) \in \mathbb{R}^{N \times M}$ helps to determine the contribution of each node for each hyperedge, and $m$ is the number of hyperedges in the hypergraph. All $\Psi (X)$, $\Lambda (X)$, and $\Omega$ are data-dependent matrices. $\Psi (X)$ is implemented by $1 \times 1$ convolution on the input features, $\Lambda (X)$ is implemented by channel-wise global average pooling followed by a $1 \times 1$ convolution as used in \cite{ref-11}, and $\Omega (X)$ is used to capture the global relationship of the features to develop better hyperedges (implemented using $s \times s$ filter, we keep $s = 7$). 
\\
\begin{figure*}
\begin{center}
\vspace*{-\baselineskip}
\includegraphics[width=1\linewidth]{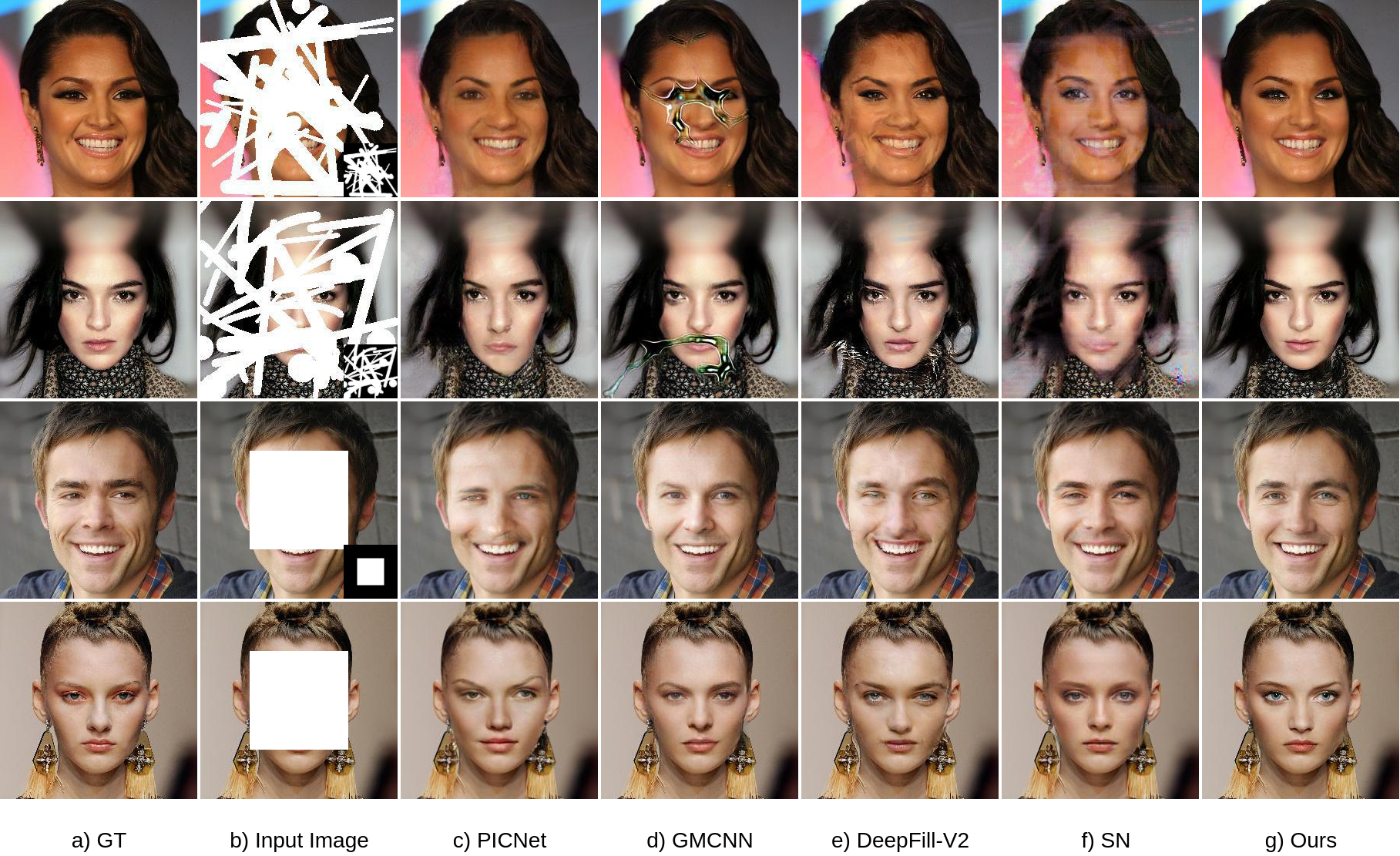}
\end{center}
  \vspace{-3mm}
  \caption{Qualitative Comparison on the CelebA-HQ dataset. From left to right $a)$ Ground Truth, $b)$ Input Image, $c)$ Pluralistic (PICNet) \cite{ref-403}, $d)$ GMCNN \cite{ref-400}, $e)$ DeepFill-V2 \cite{ref-14}, $f)$ Shift-Net (SN) \cite{ref-117}, $g)$ Ours. All the images are scaled to the size $256 \times 256$.} \vspace{-3mm}
  \label{fig:fig4}
\end{figure*}
We compute the incidence matrix $H^{l}$ as shown in the (Figure-\ref{fig:fig2}), and it can be formulated as,
\begin{equation}
    \begin{aligned}
       \mathbf {H^{\textit{l}}} &= \mathbf{\Psi} (X^{l}) \mathbf{\Lambda} (X^{l}) \mathbf{\Psi} (X^{l})^{T} \mathbf{\Omega} (X^{l}) \\
        &\mathbf{\Psi} (X^{l}) = conv (X^{l}, W_{\Psi}^{l}) \\
        &\mathbf{\Lambda} (X^{l}) = diag (conv (\hat{X}^{l}, W_{\Lambda}^{l})) \\
        &\mathbf{\Omega} (X^{l}) = conv (X^{l}, W_{\Omega}^{l})
    \end{aligned}
\end{equation}
where $\hat{X^{l}} \in \mathbb{R}^{1 \times 1 \times \hat{C}}$ is the feature map produced after global pooling of the input features, $W_{\Psi}^{l}$, $W_{\Lambda}^{l}$ and $W_{\Omega}^{l}$ are the learnable parameters for the linear embedding. To avoid the negative values in the incidence matrix $H$, which could lead to imaginary values in the degree matrices, we use the absolute values in the incidence matrix. Then we formulate our hypergraph convolution layer on spatial features as,
\begin{equation}
    \mathbf {X^{l+1}} = \sigma (\mathbf{\Delta} \mathbf{X}^{l} \mathbf{\Theta})
\end{equation}
where $\Theta \in \mathbb{R}^{C_{l} \times C_{l+1}}$ is the learnable parameters, $\sigma$ is the non-linear activation, (we use Exponential Linear Unit (ELU) \cite{ref-15}), $X^{l}$ are the input features and $X_{l+1}$ are the output features.
\subsection{Inpainting Network Architecture}
The network architecture of our proposed method is given in (Figure-\ref{fig:fig3}). We use a two-stage coarse-to-fine network architecture. The coarse network roughly fills the missing region, which is naively blended with the input image, then refine network predicts the finer results with sharp edges. In the refine network, we use the hypergraph layer with high-level feature maps to increase the receptive field of our network and obtain distant global information of the image. We use dilated convolutions \cite{ref-12} for our coarse and refine network to further expand our network’s receptive field. We also used gated convolutions \cite{ref-14} to improve our performance on an irregular mask which can be defined as,
\begin{equation}
    \begin{aligned}
        Ga&ting = Conv (W_{g}, I) \\
        Fea&tures = Conv (W_{f}, I) \\
        O = \phi& (Features) \odot \sigma (Gating) \label{eq-A12}
    \end{aligned}
\end{equation}
where $W_{g}$ and $W_{f}$ are two different learnable parameters for convolution operation, $\sigma$ is the sigmoid activation function, and $\phi$ is a non-linear activation function, such as ReLU, ELU, and LeakyReLU. We also remove batch normalization from all our convolutional layers as they can deteriorate the color coherency of the completed image \cite{ref-12}. The architecture of the discriminator used is our method similar to the PatchGAN \cite{ref-23}. We remove all the batch normalization layers and replace all the convolution layers with the gated convolution using which enforces local consistency in the completed image. We provide the discriminator with both mask and completed/original image.

\subsection{Loss Functions}
Given an input image $I_{in}$ with holes, and a binary mask $R$ ($1$ for holes), our network predicts $I_{coarse}$, and $I_{refine}$ from the coarse and refine network respectively. For the given ground truth $I_{gt}$, we train our network on the combination of losses consisting of content loss, adversarial loss, perceptual loss, and edge-loss.
\\
To force the pixel level consistency we use the $L1$ loss on both coarse $I_{coarse}$ and refine $I_{refine}$ outputs. We define content loss as,
\begin{equation}
    \begin{aligned}
        \mathcal{L}_{hole} &= || R \odot (I_{refine} - I_{gt}) ||_1 \\+ & \frac{1}{2}|| R \odot (I_{coarse} - I_{gt}) ||_1  \\
        \mathcal{L}_{valid} &= || (1 - R) \odot (I_{refine} - I_{gt}) ||_1 \\+ &\frac{1}{2}|| (1 - R) \odot (I_{coarse} - I_{gt}) ||_1
    \end{aligned}
\end{equation}
where $L_{hole}$ is the loss for the hole pixels values, and $L_{valid}$ is the loss for the non-pixels values. 
\begin{figure*}
\begin{center}
\includegraphics[width=1.0\linewidth]{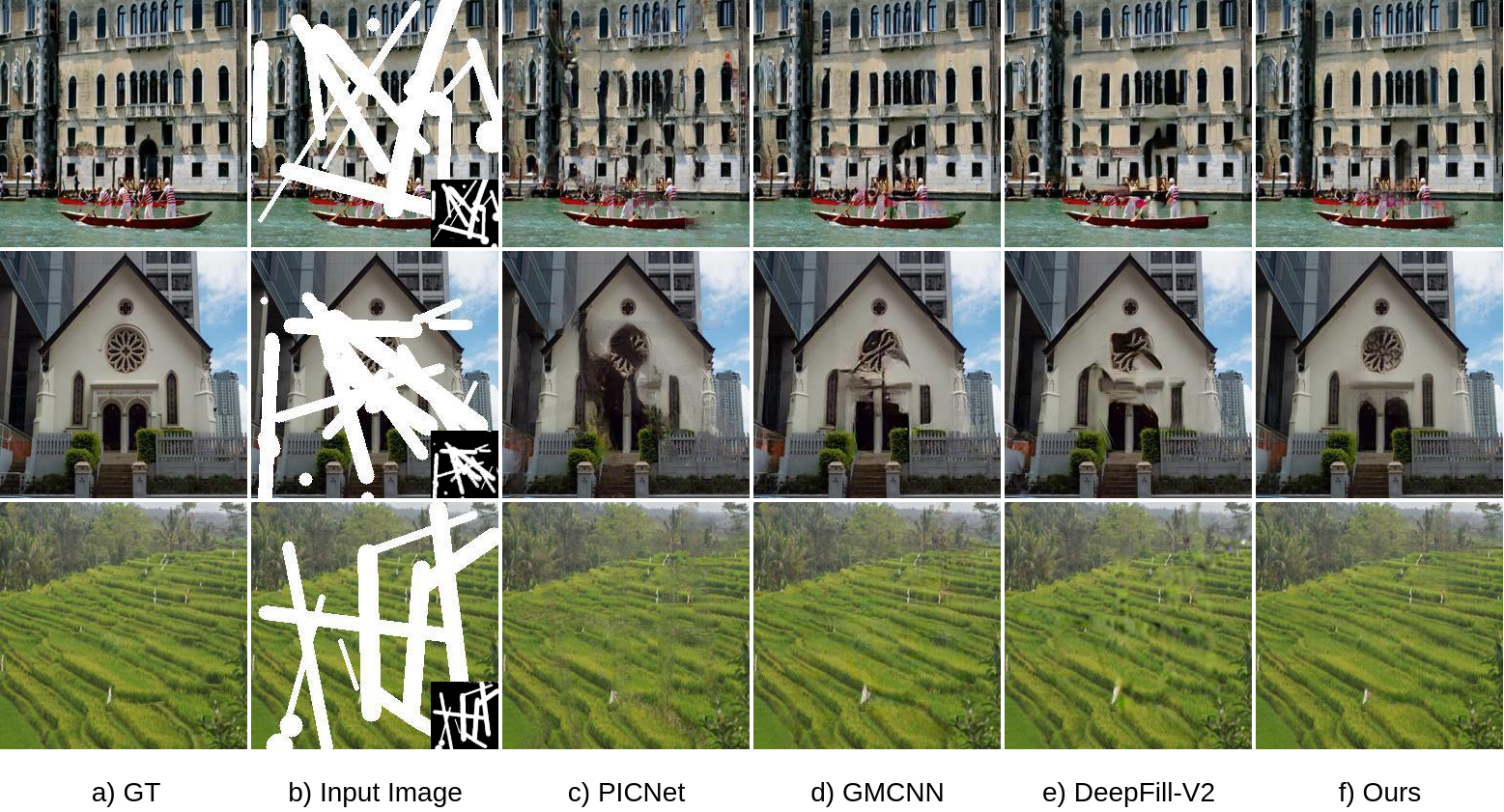}
\end{center}
\vspace{-5mm}
  \caption{Qualitative Comparison on the Places2 dataset. From left to right $a)$ Ground Truth, $b)$ Input Image, $c)$ Pluralistic (PICNet) \cite{ref-403}, $d)$ GMCNN \cite{ref-400}, $e)$ DeepFill-V2 \cite{ref-14}, $f)$ Ours. All the images are scaled to the size $256 \times 256$} \vspace{-5mm}
  \label{fig:fig5}
\end{figure*}
\\
The adversarial loss has been shown to be effective to generate realistic and globally consistent images \cite{ref-20, ref-21, ref-22}. The adversarial loss can be formulated as a min-max problem,  
\begin{equation}
    \begin{aligned}
        \mathcal{L}_{GAN} = \max_{D}\min_{G} \mathbb{E}[log (D (I_{gt}, R))] \\+ \mathbb{E} [log (1 - D(G (I_{in}), R)]
    \end{aligned}
\end{equation}
where $G$ is our image inpainting network which predicts the final completed image $I_{refine}$, and $D$ is the Discriminator.
\\
Perceptual loss has been used in many applications such as image super-resolution \cite{ref-25, ref-26}, image deblurring \cite{ref-27, ref-28}, and style transfer \cite{ref-29, ref-30}. For a given input $x$, let $\phi_{l} (x)$ denote the high-dimension features of $l^{th}$ activation layer of the pre-trained network, then the perceptual loss can be defined as,
\begin{equation}
    \mathcal{L}_{p} = \sum_{l} || \phi_{l} (G(I_{in})) - \phi_{l} (I_{gt}) ||_1  
\end{equation}
We compute perceptual loss for final prediction $I_{refine}$, and $I_{comp}$, where $I_{comp}$ is the final prediction but the non-hole pixels directly set to ground-truth \cite{ref-13}.
\\
To maintain the edges in the predicted images, we use the edge-preserving loss \cite{ref-24}. It can be defined as,
\begin{equation}
    \mathcal{L}_{edge} = ||E(I_{refine}) - E(I_{gt})||_1
\end{equation}
where $E(\cdot)$ is the sobel filter. So our total loss $\mathcal{L}_{total}$ can be written as,
\begin{equation}
    \begin{aligned}
        \mathcal{L}_{total} =& \lambda_{hole}\mathcal{L}_{hole} + \lambda_{valid} \mathcal{L}_{valid} + \lambda_{adv}\mathcal{L}_{adv} \\ &+ \lambda_{p} \mathcal{L}_{p} + \lambda_{edge} \mathcal{L}_{edge}
    \end{aligned}
\end{equation}
where $\lambda_{hole}, \lambda_{valid}, \lambda_{adv}, \lambda_{p}$ and $\lambda_{edge}$ are the weights to balance the hole, valid, adversarial, perceptual and edge loss respectively. 

\subsection{Incremental Training}
Training the deep learning approach for image inpainting on a random mask is an arduous task because of the random hole size in the testing phase. To handle this issue, we introduce a simple yet effective training technique for image inpainting. Initially, we start training our training with a very small hole percentage. Therefore initially, the network can learn to output the non-hole pixels accurately. Then gradually, we increase the hole percentage so that the network can learn a better mapping for large holes. Specifically, we train our network for K iterations and then increase the hole size. 

\section{Experiments}
\subsection{Implementation Details}
During training, we linearly scale the image’s values in the range $[0, 1]$. We trained our model on NVIDIA 1080Ti GPU with the image resolution of $256\times256$ with a batch size of $1$. We use Adams optimization \cite{ref-19} with $\beta_1 = 0.9$, and $\beta_2 = 0.999$ and the initial learning rate of $1 \times 10^{-4}$ decreasing by a factor of $0.96$ after each epoch.

\subsection{Datasets}
We evaluate our proposed network on four publicly available datasets, including CelebA-HQ \cite{ref-16}, Paris Street View \cite{ref-17}, Facades Dataset \cite{ref-404}, and ten scenes in Places2 dataset \cite{ref-18}. The CelebA-HQ dataset contains $30,000$ images of faces, we randomly sample $28,000$ images for training and $2,000$ images for testing. There are $14,900$ images for training and $100$ images testing in the Paris Street View Dataset. The Facades Datasets contains $400$ training images, $100$ validation images, and $106$ testing images(For Facades Dataset we fine-tune the model trained on Paris Street View dataset). Places365-Standard Dataset contains $1.6$ million training images from 365 scenes. We choose ten different scenes including canyon, field\_road, field-cultivated, field-urban, synagogue-outdoor, tundra, valley, canal-natural, and canal-urban. Each category contains $5,000$ training images, $100$ validation images, and $900$ testing images. We also use data augmentation, such as flipping and rotating for the Places2 and Paris Street View datasets. To evaluate our results, we train our model both on the center-fixed hole and random hole. We remove $25\%$ of the center pixels for the center-fixed hole, and for the random hole, we simulate spots, tears, scratches, and manual erasing with brushes.
\begin{figure}
\begin{center}
\includegraphics[width=1.0\linewidth]{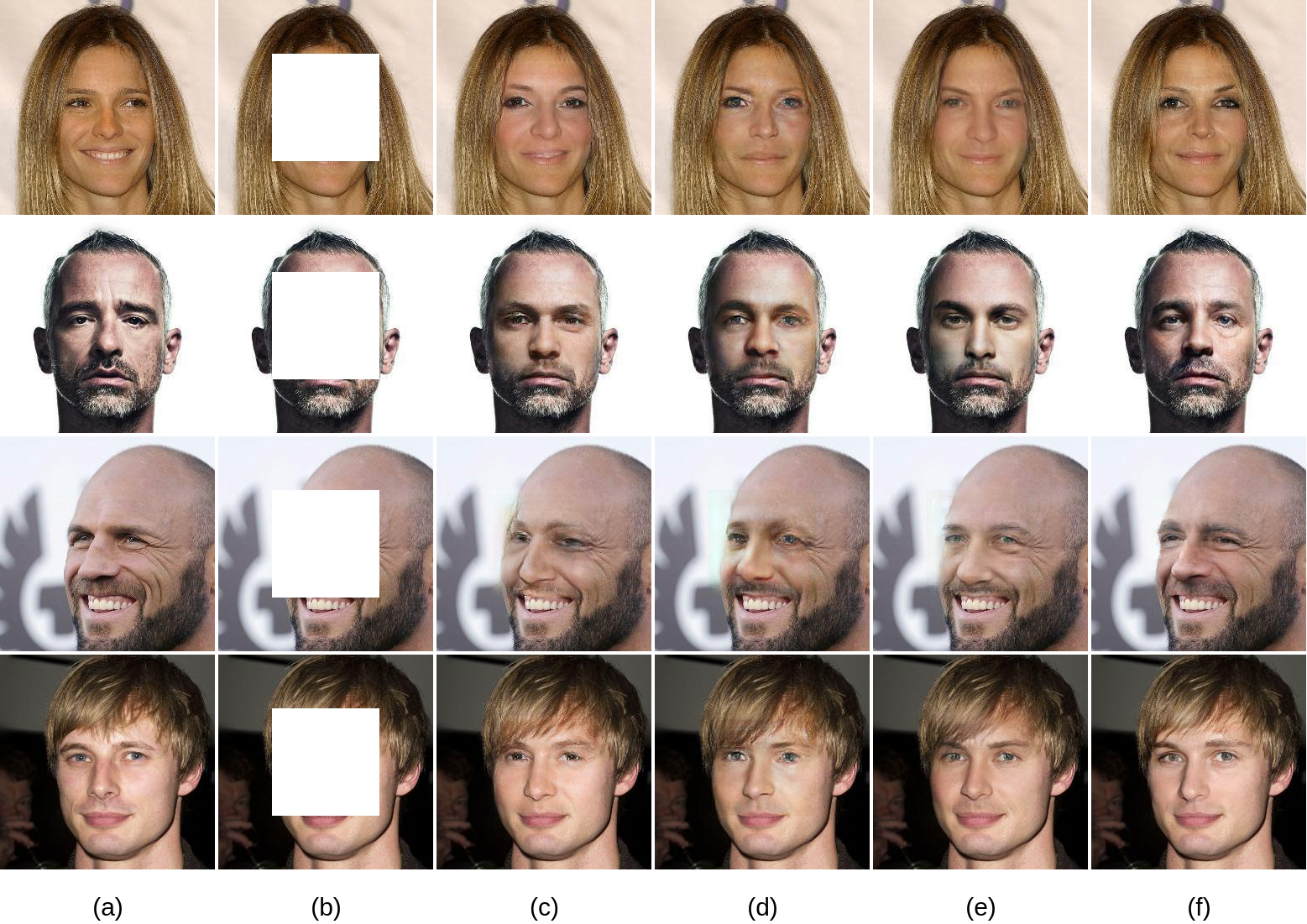}
\end{center} \vspace{-5mm}
  \caption{Comparison of different variants of our proposed method. From left to right $a)$ Ground Truth (GT), $b)$ Input Image, $c)$ w/o hypergraph attention and w/o gated convolution in discriminator, $d)$ w/o gated convolution in discriminator, $e)$ w/o hypergraph attention, and $f)$ Original method. } \vspace{-5mm}
  \label{fig:fig6}
\end{figure}
\subsection{Comparison with Existing Work}
We compare our method against the existing works, including multi-column image inpainting (GMCNN)\cite{ref-400}, DeepFill \cite{ref-14}, Pluralistic Image Inpainting (PICNet)  \cite{ref-403} (As PICNet can generate multiple results we choose the best result based on the discriminator scores for fair comparison), and Shift-Net (SN) \cite{ref-117}, for both centre masks, and irregular masks. 

\textbf{Quantitative Comparison:} As mentioned by \cite{ref-118}, there is no good numeric method for evaluating the image inpainting results because of the existence of many plausible results for the same image and mask. Nevertheless, we report, in Table-1, our evaluation results in terms of $l_1$ error, $l_2$ error, PSNR, MS-SSIM \cite{ref-401}, and Frechet Inception Distance (FID) \cite{ref-402} on the validation set of places2 and testing set CelebA-HQ datasets. In the table, we compare different approaches on random mask of different hole percentages for the both datasets. As shown in the table, our method outperforms all the existence methods in terms of $l_1$ loss, $l_2$ loss, PSNR, MS-SSIM, and FID.

\textbf{Qualitative Comparison:} Figure-\ref{fig:fig4}, and Figure-\ref{fig:fig5} shows the comparison between our method and the other existing methods on CelebA-HQ and Places2 datasets respectively. We observe that our method produce much more semantically plausible and globally consistent results even for for much larger mask region. Earlier methods performs good en enough for small mask percentage but there performance deteriorates as the mask size increases. Especially, GMCNN, and DeepFill-V2 produces severe artifacts when the hole size increases beyond $50\%$. The outputs of Shift-Net (SN) algorithm does not produce color consistent outputs. PICNet produces semantically plausible and clear results but the outputs produced are not globally consistent, this is because PICNet of the discriminator which constraints the the network to produce clearer image but looses the structural consistency of the image and hence produces artifacts in the predicted image. Our method is able to produce much more plausible and realistic outputs because of the hypergraphs, which helps the generator to learn the global context of the image, and the gated convolutions used in discriminator helps the generator learn the local contents of the image.  


\subsection{Ablation Study}
We further perform experiments on CelebA-HQ dataset to study the effects of different components of our introduced methodology. In figure-\ref{fig:fig6}, we show the comparison between different variants of our method, including $a)$ w/o hypergraph attention mechanism with normal convolution in discriminator $b)$ Replacing gated convolution with normal convolution in the discriminator, and $c)$ w/o hypergraph attention mechanism. Using normal convolution in discriminator affects the local consistency of the image and produces artifacts in the completed image. Not using hypergraph attention mechanism disturbs the global color consistency of the completed image because hypergraph convolution provides the global structure of the image.      

\section{Conclusion}
In this paper, we proposed a Hypergraph convolution based image inpainting technique, where the hypergraph incidence matrix $H$, is data-dependent and enables us to use a trainable mechanism to connect nodes using hyperedges. Using hypergraphs helps the generator to get the global context of the image. We also propose the use of gated convolution in discriminator which helps the discriminator enforce a local consistency in the image. Our proposed method produces a final image which is semantically plausible and globally consistent. Our experimental results indicate that our method gets better performance than any of the state-of-the-art methods and improves the quality of the completed image. 
Further, we can easily extend the idea of hypergraph convolution on spatial features in any other application to learn the global context of the image.

{\small
\bibliographystyle{ieee_fullname}
\bibliography{egbib}
}



\end{document}